\documentclass[]{article}
\usepackage{amsmath,amsfonts,bm}

\def\figref#1{Fig.~\ref{#1}}
\def\Figref#1{Figure~\ref{#1}}

\def\Secref#1{Section~\ref{#1}}

\def\eqref#1{Eq.~(\ref{#1})}

\def\1{\bm{1}}

\def\re{{\textnormal{e}}}

\def\rvx{{\mathbf{x}}}
\def\rvy{{\mathbf{y}}}
\def\rvz{{\mathbf{z}}}

\def\vtheta{{\bm{\theta}}}
\def\vvartheta{{\bm{\vartheta}}}
\def\va{{\bm{a}}}

\def\vf{{\bm{f}}}

\def\vk{{\bm{k}}}

\def\vq{{\bm{q}}}

\def\vv{{\bm{v}}}

\def\vx{{\bm{x}}}
\def\vy{{\bm{y}}}

\def\mK{{\bm{K}}}

\def\mQ{{\bm{Q}}}

\def\mT{{\bm{T}}}

\def\mV{{\bm{V}}}

\DeclareMathAlphabet{\mathsfit}{\encodingdefault}{\sfdefault}{m}{sl}
\SetMathAlphabet{\mathsfit}{bold}{\encodingdefault}{\sfdefault}{bx}{n}

\newcommand{\E}{\mathbb{E}}

\newcommand{\R}{\mathbb{R}}

\DeclareMathOperator*{\argmin}{arg\,min}

\title{Knowledge as Invariance - History and Perspectives of Knowledge-augmented Machine Learning}
\author{Alexander Sagel\thanks{Contact: sagel AT fortiss DOT org}, Amit Sahu, Stefan Matthes, Holger Pfeifer,\\
Tianming Qiu, Harald Rue\ss, Hao Shen, Julian W\"ormann
\\
\textit{\small fortiss GmbH - The Research Institute of the Free State of Bavaria, Germany}}
\usepackage{float}
\usepackage{graphicx}
\usepackage{easy-todo}
\begin{document}

\maketitle

\begin{abstract}
Research in machine learning is at a turning point. While supervised deep learning has conquered the field at a breathtaking pace and demonstrated the ability to solve inference problems with unprecedented accuracy, it still does not quite live up to its name if we think of learning as the process of acquiring knowledge about a subject or problem. Major weaknesses of present-day deep learning models are, for instance, their lack of adaptability to changes of environment or their incapability to perform other kinds of tasks than the one they were trained for. 
While it is still unclear how to overcome these limitations, one can observe a paradigm shift within the machine learning community, with research interests shifting away from increasing the performance of highly parameterized models to exceedingly specific tasks, and towards employing machine learning algorithms in highly diverse domains. This research question can be approached from different angles. For instance, the field of\emph{ Informed AI} investigates the problem of infusing domain knowledge into a machine learning model, by using techniques such as regularization, data augmentation or post-processing. 

On the other hand, a remarkable number of works in the recent years has focused on developing models that by themselves guarantee a certain degree of versatility and \emph{invariance} with respect to the domain or problem at hand. Thus, rather than investigating how to provide domain-specific knowledge to machine learning models, these works explore methods that equip the models with the capability of \emph{acquiring} the knowledge by themselves. This white paper provides an introduction and discussion of this emerging field in machine learning research. To this end, it reviews the role of knowledge in machine learning, and discusses its relation to the concept of invariance, before providing a literature review of the field. Additionally, it gives insight into some historical context.
\end{abstract}

\section{Introduction}
One of the most prominent researchers in deep learning, Yoshua Bengio, cites the \emph{Global Workspace Theory of Consciousness} \cite{kahneman2011} as his preferred model of human cognitive capabilities \cite{bengio2017}. According to this theory, human consciousness can be grouped into two systems, with System 1 performing intuitive, automated tasks that we can do instinctively and System 2 performing tasks that require conscious decision making and can be described verbally.

Current deep learning algorithms are particularly good at performing System 1-level tasks. For instance, if we are presented with pictures of edible plant matter and are asked to group them into nutritional categories, such as fruits, vegetables, nuts, grains, legumes, etc, we would perform this task instinctively and probably without any hesitation. The same task should also be easily accomplished by a neural network, trained with the appropriate data. Consider now an adaptation of the task, where we are asked to group the same images into botanical categories, such as leafs, fruits (in the botanical sense), roots, seeds, etc. Many of us would probably feel slightly less confident with performing this task, but after reading up the according definitions, humans would likely still perform quite well. A neural network, on the other hand, would typically require re-learning all of its parameters.

The above example exposes two remarkable cognitive capabilities present in humans that neural networks typically lack: the ability to incorporate complementary input into the task execution and the ability to generalize a System 1 level skill to changes in the problem setting. 
It makes sense to treat these two capabilities as flip sides of the same coin.  The reason can be found in the \emph{No free lunch theorem} \cite{flach2012}, since, broadly speaking, a model that is perfectly adapted to one task can not be generalized to other tasks without either forfeiting performance or infusing additional assumptions about the task or the data into it.

Unsurprisingly, the field of \emph{Informed Machine Learning} \cite{VonRueden2019} that investigates how to enhance machine learning by means of prior domain knowledge has gained considerable importance. The employed techniques include methods such as data augmentation, loss regularization, hyper-parameter design or post-process filtering of the model output \cite{VonRueden2019}. However, these approaches build upon the assumption that today's off-the-shelf deep learning models offer sufficient versatility to adapt to specified scenarios on-demand. This is unlikely the case and one of the major reasons for this has to do with the research culture of the machine learning community. As argued in \cite{chollet2019}, progress in machine learning is heavily driven by universally available and easily implementable benchmarks and improvement of a model is measured by how well it is adapted to these benchmarks. Now, what used to be a catalyst of research advances for deep learning is becoming more and more of a burden, as demand for generalization increases and adversarial attacks expose the weaknesses of highly specialized training. This bias towards specialization has been recognized and identified as a problem by the community's leading figures, such as Yoshua Bengio, Geoffrey Hinton and Yann LeCun \cite{LeCun2020}. Similar concerns were expressed by  Fran\c cois Chollet \cite{chollet2019} and Gary Marcus \cite{marcus2020}.

These debates have sparked a number of research directions that rather than studying the \emph{adaptation} of machine learning models to a specific problem or situation, focus on their \emph{adaptability.} This adaptability can refer to different aspects of the problem at hand, e.g. the lighting condition in visual data, the length of a natural-language phrase or even the skill that is to be learned itself. This white 
paper is a modest attempt to provide an overview of the most promising developments in this direction and to exemplify their relation to the concept of knowledge in AI.

\section{Knowledge as Invariance}
\subsection{From Machine Learning to Knowledge Acquisition}
The Oxford Dictionaries define knowledge as \emph{facts, information, and skills acquired by a person through experience or education} or \emph{the theoretical or practical understanding of a subject}.
Current AI systems, notably deep learning architectures, can be described as systems that acquire facts or skills through experience or education, i.e. training. Still, neural networks can be hardly considered \emph{knowledgeable} in the broader sense in which we understand this term. But what is it about knowledge that current AI methods in general and deep learning specifically fall short of? What should a "knowledgeable system" be able to do that a typical deep neural network can not?
\begin{figure}
	\begin{center}
		\includegraphics[width=0.5\textwidth]{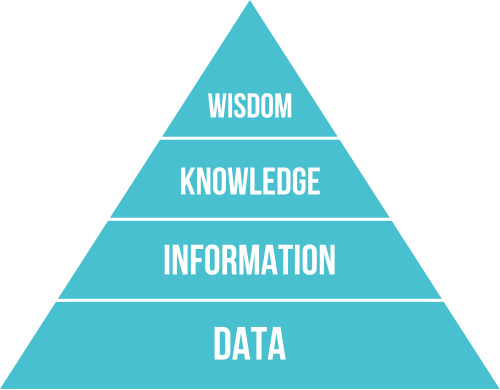}
	\end{center}
	\caption{\label{fig:dikw}The DIKW Pyramid. Source: \cite{dkiw2015}}
\end{figure} 

\Figref{fig:dikw} depicts the DIKW Pyramid \cite{rowley2007} that groups the terms \emph{data - information - knowledge - wisdom} along an abstraction hierarchy. While raw data is useless for carrying out any decisions, information infers structure and task-bound function from data by providing answers to clearly specified questions \cite{rowley2008}. 

In a way, today's established deep learning systems predominantly work on this level of abstraction. They take raw data and infer just enough rules from it, to answer questions such as "Does this image contain a cat?" or "Did this reviewer enjoy that book?". If we think of wisdom as the level corresponding to a hypothetical Artificial General Intelligence (AGI), i.e. systems that can be considered fully autonomous up to the point of asking for a higher meaning or purpose of a task, knowledge would correspond to a stage somewhere in-between the two. 

Knowledgeable machines should go beyond answering well-defined task-specific questions and see a slightly bigger picture without necessarily becoming fully autonomous in that. Davenport and Prusak \cite{davenport2000} describe knowledge as
\begin{quote}
	\textit{[...] a fluid mix of framed experience, values, contextual information, expert insight and grounded intuition that provides an environment and framework for evaluating and incorporating new experiences and information.}
\end{quote}
Note how this definition puts emphasis on adaptability. This coincides with the widely accepted notion that only by solving \emph{transfer tasks}, we can verify that we have knowledge in a field. \emph{Informed} systems differ from \emph{knowledgeable} ones in how rigid they are with respect to a skill or a situation.  The transition from informed to knowledge-based AI thus heavily depends on how invariant they become. 

Admittedly, machine learning has 
always been about invariance. Deep convolutional neural nets, for instance, learn low-level invariances on the pixel level, e.g. invariance to translation operations. However, invariance at higher levels of abstraction is still rare to find in contemporary machine learning models. In particular, we believe that the following three types of invariance are crucial for knowledge-based AI models.

\subsubsection{Invariance to the Skill}
\label{sub:skill}
Deep Learning owes much of its success to the \emph{supervised learning} paradigm. At the same time, the emphasis on supervised learning has been identified as one of the major limiting factors in achieving AGI \cite{LeCun2020}. Formally, most of the common supervised learning problems can be written as some form of function approximation, in which the task is to find a $\vtheta$-parameterized function  
\begin{equation}
f_\vtheta:\mathcal X\to \mathcal Y
\end{equation}
that maps (approximately) each input data sample $\vx$ to a label $\vy$ according to the rules that have been learned from a set of labeled training data
\begin{equation}
\{(\vx_i, \vy_i)\}_{i=1,\dots,N}.
\end{equation}
We generally assume that the training set has been independently sampled from a joint distribution $p(\rvx, \rvy)$ such that the function approximation can be phrased as likelihood optimization for the trainable parameter $\vtheta$, e.g. the entirety of weights in the neural network to be trained. It should go without saying, that by training a model in such a way, it does not acquire knowledge regarding the data it has been fed with, as knowledge is characterized by being transferable from one skill to another.

In order to acquire knowledge, a model should not just yield one particular function, but provide a possibility to infer different kinds of functions with respect to the data it has been trained on. For instance, a model that has been fed with natural images should not just be able to classify the images in categories, but perform semantically related tasks, such as detecting which kinds of objects tend to appear close to each other, how they are spatially aligned with respect to each other and so on. 

\subsubsection{Invariance to the Data Distribution}
Most common optimization objectives in machine learning are derived from likelihood maximization. This builds upon the implicit assumption that training and test data are sampled from the same distribution. In reality, data is gathered under conditions that can change. Arjovski et al. provide an illustrating example of this issue in  \cite{arjovsky2019}: consider a neural network that is trained to classify images into photographs of cows and camels. As such pictures are usually taken in the environments one would expect to find the respective ruminants, the network will likely tend to assign green grassland to the former, and sandy landscapes to the latter category. While from a statistical point of view, this is a perfectly legitimate way to infer correlations from the data, it contradicts another aspect of knowledge that transcends  the System I-level skillset, namely the capability of adapting to changes in situation, context or environment.

This capability is typically referred to as \emph{out of distribution} (OOD) generalization. Unfortunately, there is no universally accepted metric that evaluates the OOD capabilities of a model, since by its very definition it does not characterize machine learning models by what the generalize to, but by what they do not generalize to (the training data distribution).

As an illustration of the difficulty to define an appropriate OOD objective, consider again the supervised learning task of identifying the function
\begin{equation}
f_\vtheta:\mathcal X \to \mathcal{Y}, \ \vtheta\in\R^p
\end{equation}
that maps from a data sample $\vx\in \mathcal X$ to a label $\vy\in \mathcal Y$, where the joint distribution $p_{e}(\vx, \vy)$ is determined by the \emph{environment} $ e \in \mathcal{E}$. Assuming an $\ell_2$ loss, we could define the environment-dependent objective function $L^e$ as
\begin{equation}
L^e(\vtheta) = \E_{\rvx, \rvy\sim} p_e[\|f_\vtheta(\rvx)-\rvy\|^2].
\end{equation}
But in order to minimize such an objective for all environments $ e \in \mathcal{E}$ fairly, we would need to make assumptions about the statistics of the environments themselves. For instance, the obvious optimization problem
\begin{equation}
\min_{\vtheta}\E_{\re\sim p(\re)}[L^\re(\vtheta)]
\label{eq:odd_obj}
\end{equation}
requires a model for the distribution $p(\re)$ of environments. Note that estimating $p(\re)$ from  data would require the  training data to be representative of the environments to appear in testing phase, which contradicts the premise of OOD.
One such possible assumption is that the function can be written as
\begin{equation}
f_\vtheta = \varphi_{\vtheta}\circ\phi,
\end{equation}
where $\phi$ has the property that
\begin{equation}
\hat{\vtheta} = \argmin_\vtheta L^e(\vtheta)
\end{equation}
does not depend on $e$. The function $\phi$ is then said to \emph{elicit} an $\mathcal E$-invariant predictor \cite{arjovsky2019}. For instance, in the cow/camel example above, a function that removes the background from an image and replaces it by neutral pixels, elicits an invariant predictor of the class. Again, it is important to be aware of the implicit assumptions that we make about $\mathcal E$. In this case, we assume that a change of $e$ mainly effects the background of an image. If $\phi$ is trained from the data, then we need to trust that if it elicits an invariant predictor for the environments in $\mathcal E$, it does so also for all possible environments that could appear in the testing phase.

A slightly different attempt to formalize OOD has been given in \cite{Greenfeld2019}. The authors assume a joint train probability $p_\mathrm{source}(\rvx, \rvy)$ between data $\rvx$ and label $\rvy$ that does not necessarily coincide with the unknown target distribution $p_\mathrm{target}(\rvx, \rvy)$. However, it is assumed that the conditional distribution of $\rvy$, given a realization of $\rvx$, is universal to all possible conditions, i.e.
\begin{equation}
p_\mathrm{source}(\rvy|\rvx=\vx)=p_\mathrm{target}(\rvy|\rvx=\vx)
\end{equation}
for all possible $\vx\in\mathcal X$ and no matter what the target environment looks like. The authors approach this aim by learning the function $f_\vtheta:\mathcal X\to \mathcal Y$ such that the property of $\rvz:=\rvy - f_\vtheta(\vx)$ being independent $\rvx$ is fulfilled. To this end a measure of independence that can be easily optimized is necessary.

\subsubsection{Invariance to the Data Syntax}
Data is always redundant and this is true regardless of it being present in the form of visual data, audio, text or any other possible data modality. This redundancy manifests in a set of implicit rules that determine how meaning is encoded within the respective data modality. For documents written in natural language, for instance, these rules are determined by grammar, whereas in natural images these rules stem from physical laws, as well as our geometrical notion of objects, backgrounds, compositions, etc. When we as humans process data that we gather through sensory stimuli, we are able to abstract knowledge from its syntactical configuration.

Deep learning has become so successful precisely because CNNs are capable of learning representations that are invariant to syntactical clutter such as the spatial location or the scale of an object in an image. However, this degree of invariance in neural networks is mostly limited to visual and sequential data, and specifically
not complex, structured or compositional data types used in knowledge representation, such as
\begin{itemize}
	\item Tables,
	\item Graphs,
	\item Algebraic or Logical Expressions,
	\item Complex Natural-Language Phrases,
	\item Sets.
\end{itemize}
Indeed, neural networks are pure vector processing machines. This means that they realize mathematical functions that map from one real-valued, finite vector space to another. The involved sets are thus naturally equipped with mathematical structure that symbolic data, to name an example, does not possess. Specifically, euclidean vector spaces have a metric, and hence allow for distance-based classification. Functions on vector spaces can also be equipped with properties such as linearity or differentiability, enabling optimization via gradient descent. Non-vector data, on the other hand, lacks many of the conveniences that neural networks relies on, as listed in the following.

\textit{Lack of interface for non-euclidean data structures.} Neural networks expect vectors with a fixed dimension at its input and output. While down- and upscaling is possible for input images of different resolutions, normalizing the size of symbolic data is in general not trivial. Moreover, vectors induces a natural order of their elements. This can also not be guaranteed for non-euclidean data, such as graphs or sets.

\textit{Lack of a universally applicable inductive bias.} Consider the task of continuing the number sequence $1,2,3,4\dots$ Humans that learn to count from an early age on will naturally make the assumption that this is the beginning of the sequence of natural numbers.
Neural networks are universal function approximators. This implies that without any additional assumption about the data at hand, there is no reason to believe that a neural network would continue the sequence the same manner as us.  These kinds of prior assumption about the data is called the \emph{inductive bias}. Deep learning has an inductive bias towards translation invariance and self-similarity due to the weight-sharing in convolutional filters.
However, these assumptions do not hold for non-image data, in general. The limitations of the inductive bias in deep learning have been recently illustrated by the benchmark presented in \cite{chollet2019}. It contains seemingly simple, low-dimensional abstract toy examples that are easily solved by humans within minutes but are particularly difficult to learned by means of neural nets.

\textit{Lack of continuity.} Besides the fact that conversion from continuous to discrete data and back are non-trivial and leads to loss of information, the absence of continuity makes it almost impossible to perform gradient-based optimization as it is common in deep learning. 

Real syntactical invariance thus requires methods to efficiently process non-euclidean data. Unlike distribution and skill invariance, that demand redefining the learning procedure, syntactical invariance can be realized on an architectural level, e.g. by means of neural layers that generalize convolutions to non-euclidean data. 

\subsection{Scope of This Work}
This work provides a survey of developments in machine learning that can be characterized as approaches to increase invariance towards the three aspects of AI problems discussed in the previous subsection. After providing some historical context on symbolic knowledge, it discusses recent developments in deep learning.

First, different architectural elements are reviewed that can be used to increase syntax invariance for different data modalities.
In particular, we consider \emph{attention} mechanisms that aim at extracting the relevant fragment within the incoming data and thus reducing the sensitivity to syntactically redundant input. Furthermore, \emph{capsule}-based neural networks are investigated as a mean to disentangle visual entities from their geometric configuration.

We then proceed to review approaches that generalize neural networks to non-vector data, in particular graphs and sets, mentioning also results from recent studies o \emph{group action symmetries} that are crucial for theoretical understanding about inductive biases in CNN-like structures.

Finally two learning paradigms, namely \emph{meta-Learning} and \emph{self-supervised learning}, which have attracted increasing interest during the recent years and have the potential to enhance skill and distribution invariance, are reviewed. Additionally, we take a look at \emph{metric learning} that has similar potential.

\subsection{Relation to Informed Machine Learning}
Knowledge is a subject of interest across many fields within machine learning. 
In this section, we would like to take a closer look  at the field of \emph{Informed machine learning}, described by von R\"uden et al. in \cite{VonRueden2019}. The motivation is that it can be considered complimentary to the direction emphasized in thus work. Speaking in broad terms, Informed machine learning focuses on questions like "How do I restrict my video prediction model to learning only physically possible scenarios?" or "How do 
I tell my autonomous driving system that the traffic conditions have changed?" In short, it studies approaches that adapt general-purpose machine learning models to the problem-specific conditions by incorporating domain knowledge. 

A major difference between \cite{VonRueden2019} and the present work is in the definition and characterization of the term \emph{knowledge}. While this work agrees with \cite{VonRueden2019} about how knowledge relates to \emph{information} with regards to its level of abstraction, we emphasize the fluidity and adaptability of it. In other words, our emphasis is on the fact that knowledge, once gained, can not only be applied to one particular, rigid problem setting, but adapts to the peculiarity of each given situation. By contrast, von R\"uden et al. stress the aspect of formalization. After defining knowledge as \emph{validated information}, the authors explain how it is characterized by the degree of formalization with regards to its representation. This generally implies that knowledge can be expressed using natural language or a similar system of communication.

As a consequence, von R\"uden et al. treat knowledge as input that is usually provided to the system from an external source, typically by an expert who gained his knowledge from domain experience inaccessible to the model and with the capability to formalize it in a machine-readable way. This input is supposed to enhance machine learning models by additional, formalized insights that could not be incorporated within the training phase.

We, on the other hand, do not treat knowledge as external to the system but as something it acquires from possibly heterogeneous input and can be applied in the context of different scenarios. Therefore, rather than investigating techniques that leverage external, formalized  inputs within machine learning models, we focus on architectural devices and learning paradigms that permit us to learn models in a way such that adaptation to new, unseen scenarios is carried out as seemlessly as possible. 

As an example, consider again the camel/cow classification problem from before. Recall that the problem consists of building a classification system that tends to wrongly include the semantically irrelevant scenery into the class assignment.  Looking at the problem from the point of view of Informed machine learning, we would ask ourselves how to appropriately formalize these changes in scenery and how to communicate them to the model. By contrast, from the perspective of invariance, we are more interested in training the  model in such a way that the background is not taken into account when the classification is performed.

That is not to say that we expect invariance-based knowledge to supersede Informed machine learning at some point. On the contrary, we expect that these two fields will complement one another in the future. It is thus important to be aware of the difference in how knowledge is defined and characterized in these two related, but distinct research fields.

\section{Neural Symbolic Integration}
As stated in the previous section, the transition from informed to knowledge-based AI heavily depends on how invariant they become. 
Logical reasoning can provide mathematically sound invariance to tricky situations. For example, symbolic logic has been used to define properties over the whole system using formal methods over the software. This ensured that the software is safe against known risky behaviours.
Symbolic reasoning inherently builds upon invariant properties that are mathematically true.
Thus, for AI to be knowledgeable it needs absolute invariance to show common sense (trivial situations) and expert behaviour as well.

Early approaches in reaching invariance hence relied on integration of the connectionist approaches with symbolic reasoning. However, in recent years and in line with the success of deep learning, research emphasis has shifted towards achieving invariance by means of design choices in the connectionnist system itself, without relying on additional guidance from symbolic AI. This section provides an overview of techniques that offer invariance via \emph{Neuro-Symbolic Integration} (NSI), before we can dive into the recent, purely connectionnist approaches later on.

Traditionally, an artificial neural network (ANN) was understood as a connectionist system that acquired expert knowledge about the problem domain after training (invariance to the skill). ANNs required raw data and were able to generalize to unencountered situations. However, the obtained knowledge was hidden within the acquired network architecture and connection weights. Symbolic systems, on the other hand, utilized complex and often recursive interdependencies between symbolically represented pieces of knowledge (invariance to the data distribution). Realizing the machine learning bottlenecks of using any of these paradigms in isolation, integrated Neuro-Symbolic systems were proposed. These hybrid systems were expected to combine the two invariances--skill and data distribution--to make the combined system robust to both.

\begin{figure}[H]
\center{\includegraphics[width=\textwidth]
{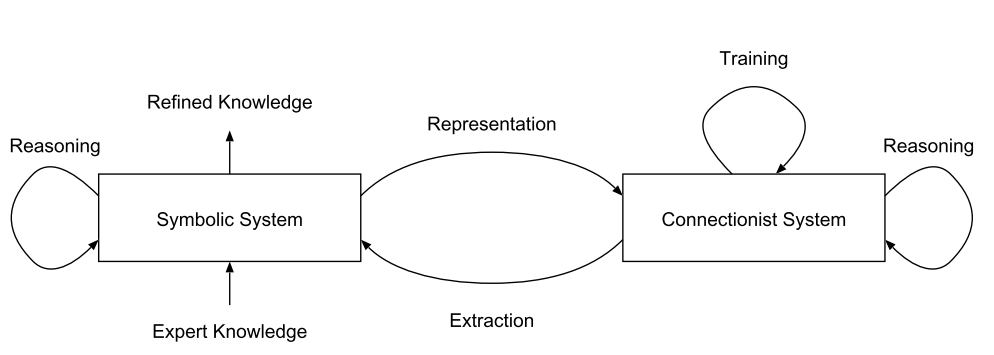}}
\caption{\label{fig:ns_learn_cycle} Neuro-Symbolic learning cycle. Source: \cite{bader2005}}
\end{figure}

Earlier methods of Neuro-Symbolic systems addressed the Neuro-Symbolic learning cycle as depicted in Figure \ref{fig:ns_learn_cycle} \cite{bader2005}. A front-end (symbolic system) fed symbolic (partial) expert knowledge to a connectionist system (ANN) that possibly utilized the internally represented symbolic knowledge during the learning phase (training). Knowledge extracted after the learning phase was fed back to the symbolic system for further processing (reasoning) in symbolic form. 

Later, \cite{garcez2009} described the Neuro-Symbolic system as a framework, where ANNs provide the machinery for parallel computation and robust learning (invariance to noise), while symbolic logic provides an explanation of the network models. These explanations facilitate the interaction with the world and other systems (invariance to skill). It is a tightly-coupled hybrid system that is continuous (ANN) but has a clear discrete interpretation (logic) at various levels of abstraction. These were able to extract logical expressions from trained neural networks and used this extracted knowledge to seed learning in further tasks. In other words, neural networks were used to tackle the invariance to the noisy data and the symbolic logic was used to obtain the invariance to skill. And as a hybrid model, it was expected to solve knowledge-based tasks. 

In the last decade, Neuro-Symbolic Integration faced many challenges and contributions \cite{garcez2015}. Prominent yet not fully solved challenges are as follows:
\begin{itemize}
\item Mechanisms of structure learning: Symbolic logic like hypothesis search at concept level (ILP) vs statistical AI using iterative adaptation processes.
\item The learning of generalization of symbolic rules
\item Effective knowledge extraction from large-scale networks for purposes like explanation, lifelong learning, and transfer learning
\end{itemize}

In contrast to early approaches using first order logic, there was a shift towards using non-classical logics \cite{besold2017}  e.g. Temporal Logic \cite{pnueli1977}, Modal Logic \cite{garcez2007}, Intuitionistic Logic \cite{dalen2002}, Description Logic \cite{krotschz2013}, and logic of intermediate expressiveness e.g. Description Logic \cite{krotschz2013},  Inductive Logic Programmming using propositionalization methods \cite{blockeel2011}, Answer-Set Programming \cite{lifschitz2002}, Modal logic \cite{garcez2007} or Propositional Dynamic Logic \cite{harel2001}.

Traditionally, Neuro-Symbolic integration was employed to integrate cognitive abilities (like induction, deduction, and abduction) with the brain method of making mental models. Computationally, it addressed the integration of logic, probabilities, and learning. This led to the development of new models with the objective of robust learning (invariance to data distribution) and efficient reasoning (invariance to skill). Some success was achieved in various domains like simulation, bioinformatics, fault diagnosis, software engineering, model checking, visual information processing, and fraud prevention~\cite{penning2010, penning2011, garcez1999}.

In parallel, another approach was using methods like probabilistic programming~\cite{gordon2014} for the generative ML algorithms like Bayesian ML. Probabilistic programs are functional or imperative programs with two additional abilities: (1) obtain values at random from distributions, and (2) condition values of variables via observations. This allows probabilistic programming to understand the program's statistical behaviour. They can also be used to represent \textit{probabilistic graphical models}~\cite{koller2009} which in turn are widely used in statistics and machine learning. These models have diverse application areas like information extraction, speech recognition, computer vision, coding theory, biology and reliability analysis.

\section{Recent Developments}
\subsection{Innovations in Neural Network Architecture}

\subsubsection{Attention}
Human beings can focus on a specific area in the field of view or recent memories to avoid over-consuming energies.  
Inspired by the visual attention of human beings, the attention mechanism in deep learning is a concept that summarizes approaches to extract the most informative subsets from sets of data. It can aid in distilling the essential content from data, making the model thus more invariant to how the data is organized syntactically.

Attention has risen to popularity in neural machine translation (NMT). 
Many classical NMT approaches are based on an encoder-decoder architecture, where the encoder maps phrases word by word to hidden state vectors and the decoder is trained to model the probability of phrases in the output languages conditioned over these vectors. 
Both the encoder and the decoder are typically realized in a recurrent manner. 
The translation is then formulated as a likelihood maximization given the probabilistic language model of the decoder and the conditioning over the hidden states.

One problem with this approach is that it does not account for sentences of different lengths. 
In long sentences, the semantic context for each word spreads out differently from shorter sentences. 
To account for this, the authors of \cite{bahdanau2014} have proposed to include an attention mechanism that maps a subset of the hidden state vectors in an encoded sentence to a fixed-length \emph{context} vector which is then fed to the decoder instead of inputting the hidden state vectors directly. The attention is implemented as a weighted sum of normalized exponential functions.

In a work on transformer architectures \cite{vaswani2017}, attention is defined more formally as a function
\begin{equation}
\begin{split}
\R^{d_k\times n_q}\times\R^{d_k\times n_k}\times\R^{d_v\times n_k} &\to \R^{d_v\times n_q}, \\
\mQ, \mK, \mV &\mapsto \mathrm{Attention}(\vq, \vk, \vv).
\end{split}
\label{eq:attention}
\end{equation}
The columns of $\mQ, \mK, \mV$ are called \emph{queries}, \emph{keys} and \emph{values}, respectively. This function computes for each query $\vq_i$ an attention vector $\va_i$ by returning a weighted sum of all values, i.e.
\begin{equation}
\va_i = \sum_{j=0}^{n_k}\alpha_{i,j} \vv_j.
\end{equation}
The weights are determined from some measure of similarity between the queries and keys. A weight $\alpha_{i,j}$ is high, if $\vq_i$ and $\vk_j$ are similar according to the measure, and close to $0$ otherwise. In neural networks, attention blocks can be used in different ways. A fixed number of query vectors could be implemented as trainable parameters of the attention blocks that receive sets of key-value pairs as an input and return an attention value as an output.

The query-key-value function in \eqref{eq:attention} gives rise to  three categories of attention mechanisms, namely \emph{spatial attention}, \emph{self-attention} and \emph{channel-wise attention}. 

\emph{Spatial attention} imitates human visual attention in a way such that the network is able to focus on significant semantic areas in input images for final decision makings. 
Queries describe ultimate classification or detection results, while values  describe pixel-level image areas and keys are feature maps extracted by convolutional neural networks (CNN).
For instance, the image  captioning work \cite{Xu:2015:SAT:3045118.3045336} is based on a similar principle as the NMT approach \cite{bahdanau2014} described above, but the attention layer is used to extract important regions from the input image, rather than phrases from a sentence.  Many works use spatial attention concepts to improve detection performance or enhance interpretability \cite{kim2017interpretable,pang2019mask}.

Another important special case of attention in the sense of \eqref{eq:attention} is \emph{self-attention}, in which $\mQ=\mK=\mV$ holds. Self-attention computes a representation of an input tuple of feature vectors based on their similarity between each other \cite{vaswani2017,ramachandran2019stand}. This concept resembles the non-local mean in image processing.   \cite{Wang2018}.

\emph{Channel-wise attention} is used predominantly in computer vision tasks by weighing channels of convolutional layers.
Similar to the aforementioned spatial attention, queries describe final classification or detection outputs, while keys and values are feature outputs, extracted from each channel of convolutional layers, because the channels are known to be activated by specific image patterns. 
For example, the work \cite{zhang2018} considers CNN channel features in a pedestrian detection task and observes that different channels respond to different body parts.
 
An attention mechanism across channels is employed to represent various occlusion patterns in one single model, such that each pattern corresponds to a combination of body parts. 
The adjusted occlusion features $\vf_{occ}$ can be written as
\begin{equation} \label{eq:occ}
    \vf_{occ} = \mathbf{\Omega}^T \vf_{chn},
\end{equation}
where $\mathbf{\Omega}$ represents the weights on channel features $\vf_{chn}$. Likewise, the work \cite{hu2018squeeze} uses channel-wise attention to aggregate the information from the entire receptive field. 

\subsubsection{Capsules}
The original motivation behind \emph{capsules} was to disentangle visual entities and their geometric relation to each other. Early works described capsules as neural modules consisting of \emph{recogniton} and \emph{generation} units. Both kinds of units are realized via hidden convolutional layers.
\begin{figure}
	\center{\includegraphics[width=0.8\textwidth]
		{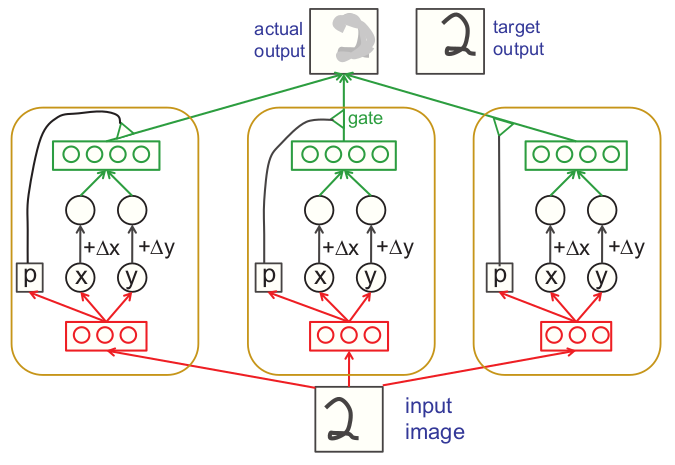}}
	\caption{\label{fig:capsule}Depiction of a capsule layer. Source: \cite{hinton2011}.}
\end{figure}
\figref{fig:capsule} depicts a capsule layer as described in \cite{hinton2011}. In the \emph{transforming autoencoders} introduced by that work, the recognition units process the image at its input and returns two parameters, a probability $p$ that a particular entity is present in the picture, and a vector $\mT$ of pose coordinates (\figref{fig:capsule}: $\mT=\begin{bmatrix}x & y\end{bmatrix}^\top$). These values are passed on to the generation units, along with a vector $\Delta\mT$ that describes the change in pose  (\figref{fig:capsule}: $\mT=\begin{bmatrix}\Delta x & \Delta y\end{bmatrix}^\top$). As a result, the generation units create a new image from the visual entities and the new poses created from $\mT$ and $\Delta \mT$. The contribution of each capsule to the generated output is determined by the presence probability $p$.

It is known that the features extracted by convolutional neural networks become more complex and expressive with increasing number of layers \cite{lee2009}. This due to the translationally equivariant nature of convolutional filters as well as the fact that different semantic features appear in different constellations throughout the training data. This is also the case for capsule neural networks but the effect is reinforced by the additional pose information provided during the training process.

As an illustrating example, assume that we want to train a transforming autoencoder with images of faces under different pose transformations. This kind of data is typically available and readily labeled in publicly accessible datasets. During training, the output image would contain the result of the change in pose. Different facial features such as mouth, ears, eyes or nose will behave differently under a given pose transformation and thus be captured by different capsules. The contribution of each capsule to the generation at the output of the transforming autoencoder is determined by $p$. If the facial feature modeled by a capsule is absent from the image $p$ should be close to $0$.

A capsule layer can be thus viewed as an architectural device to decompose the input into its semantic components. Several capsule layers can be stacked to capture features of increasing complexity.

While capsule neural networks, like convolutional nets are by design an instrument for visual data, it is interesting how they incorporate different data modalities into the learning process. Recent capsule architectures are capable of combining euclidean with symbolic representations. For instance, the \emph{Stacked Capsule Autoencoder} \cite{kosiorek2019} decomposes images into sets of objects and then uses set processing methods to organize the sets into constellations of objects.

\subsection{Invariance in Different Data Types}
\subsubsection{Neural Set Processing}
Sets are one of the most fundamental non-euclidean data types. They appear in classical combinatorial problems that are known from theoretical computer science, but also in fields like computer vision in the form of, for instance, point clouds. As such, they occupy a special position within non-euclidean data as essentially all other practically relevant data types can be derived from sets by adding additional structure. For instance, a word is a set of letters equipped with an order, a graph is a set of nodes equipped with a pairwise relational structure.  Making neural networks capable of dealing with sets thus potentially renders them applicable to all kinds of data that can be derived from a set.

Unlike elements of a vector space, sets can be of different cardinalities and do not have a natural order. To process sets, a neural network should thus be able to handle inputs of different sizes and be invariant to permutations. This permutation invariance is an elimentary inductive bias in set processing and must be considered both when neural networks process sets as their input or return sets as their output.

Since recurrent neural networks (RNNs) can handle sequences of different lengths, they have also been employed to process sets. In order to achieve permutation invariance, attention has been used. For instance, the work \cite{vinyals2015} describes a system where an LSTM generates queries to compute attention vectors from sets. The technique is employed for combinatorial tasks such as sorting. However, purely feed-forward structures have also been used for set processing, e.g. by generalizing the convolution operation to sets \cite{li2018a}.

An important theoretical result on permutation invariance has been provided in \cite{zaheer2017}. Given a function
\begin{equation}
\begin{split}
f:\mathcal{X}&\to\R,\\
X&\mapsto f(X),
\end{split}
\end{equation}
where $\mathcal{X}$ is the set of sets containing elements of a countable set $\mathfrak X$, it can be shown that $f$ is invariant to permutations of its argument, iff it can be written as
\begin{equation}
f(X)=\rho \left(\sum_{\vx\in X}\phi(\vx)\right),
\label{eq:f_perm_invariant}
\end{equation}
with $\rho:\R\to \R$ and $\phi:X\to\R$ being appropriate transformations. This result provides an easy-to-implement guideline in designing models for set processing.

Generally, attention is a popular approach in handling sets. The reason is that attention modules can be used to implement functions of the form in \eqref{eq:f_perm_invariant} \cite{lee2018}.

\subsubsection{Graph Neural Networks}
Like sets, graphs generalizes euclidean data types and at the same type can be used to describe a variety of knowledge representations, such as social networks, multi-view images or molecule structures. The survey \cite{wu2020} provides a comprehensive overview over the recent trends in Graph Neural Nets (GNN). 
A GNN can refer both to an \emph{intra-graph} framework, that operates on a node or edge level, e.g. for segmenting a graph into semantically distinct clusters, as well as an inter-graph framework, that, for example, performs classification of adjacency matrices.
Overall, intra-graph frameworks are less common. A noteworthy example is \cite{kipf2016} that presents a semi-supervised classification architecture that operates on partially labeled graph nodes.

Graph neural networks have been realized both by  recurrent and feed-forward architectures.
\begin{itemize}
	\item \emph{Recurrent} GNNs typically process each node by a recurrent unit such as a Long-term Short Memory (LSTM) or a Gate Recurrent Unit (GRU). 
	Each unit receives inputs from the units corresponding to its neighboring nodes. Works in this category, such as \cite{dai2018} or \cite{li2015} belong to the pioneering approaches to of GNNs \cite{wu2020}.
	\item \emph{Convolutional} GNNs aim at generalizing the concept of convolutions to from signals defined on regular grids to signals on graphs. 2D images, for instance, can be viewed as a special case of graphs, where each pixel is described by a node and the neighboring pixels constitute the neighborhood of adjacent nodes. Graph convolutions, like regular ones, can be carried out in the spatial \cite{monti2017,xu2018, velivckovic2018} and the spectral domain \cite{bruna2013,li2018,kipf2016}, by applying an appropriate transform of the graph data. One important question of ongoing research in the context of Convolutional GNNs is the design of appropriate pooling layers \cite{diehl2019}.
	\item Similarly, \emph{attention} based mechanisms have also been employed \cite{velivckovic2017}.
\end{itemize}

GNNs have been widely applied to non-supervised learning tasks, such as graph embedding \cite{perozzi2014} and graph generation \cite{bojchevski2018}.

\subsubsection{Group Action Symmetries}
As we have seen, considerable effort is put into generalizing convolutions to data structures such as graphs, sets or manifolds. This is not without a reason. Weight sharing in convolutional layers of deep networks without doubt provides a strong prior for the most common deep learning applications \cite{ulyanov2018}. While it is not entirely understood what exactly it is about convolutional neural nets that make them capture the essential information from visual data, robustness towards certain transformations such as translations of images seem to play an important role in it \cite{Mallat2016}. Most prominently, convolutions are equivariant to spatial translations, which is advantageous for visual data, as translations typically have little impact on the semantic content of an image. 

However, if we want to apply deep learning as successfully to non-image inputs, we need to generalize these kinds of symmetries to more exotic types of data which turns out to be tricky. Nevertheless, some theoretical results on this matter have been presented in \cite{ravanbakhsh2017}, with regards to how parameter sharing induces equivariances with respect to some exemplary group operations on the input, such as rotations and permutations. Later, the work \cite{kondor2018} has claimed some stronger results, by showing that a convolutional structure is not only  a sufficient, but also a necessary condition for equivariance with respect to certain group actions.

\subsection{Recent Trends in Learning Paradigms}
\subsubsection{Meta-Learning}
\emph{Meta-Learning} refers to a class of approaches in predominantly supervised learning settings that can be vaguely described as "learning to learn" \cite{finn2017}. Traditional supervised learning problems are typically formulated in terms of \emph{training data} and \emph{test data}, where the training data is used to optimize a parameterized function for classifying or regressing test data samples that are assumed to be sufficiently similar to the training data in terms of labeling and statistics. By contrast, in meta-learning, once a model has been trained, it is not directly used to predict labels of unseen data samples, but rather to once again learn the prediction on a small unseen data-set. This field of study has considerable overlap with few-shot classification \cite{yoon2019, vinyals2016}.

Meta-learning problems are often framed in terms of \emph{support sets} and \emph{query sets}. A training set
\begin{equation}
\mathcal{X}_\mathrm{train}=\{(\mathcal S_1, \mathcal Q_1),\dots,(\mathcal S_{N_\mathrm{train}}, \mathcal Q_{N_\mathrm{train}})\}
\end{equation}
contains $N_\mathrm{train}$ pairs of support and vector sets. A meta-learning framework uses $\mathcal{X}_\mathrm{train}$ to generate a deep learning model that can be easily trained on the support set of a new, unseen pair $(\mathcal{S}_\mathrm{test},\mathcal{Q}_\mathrm{test})$, such that it generalizes to the samples in $\mathcal{Q}_\mathrm{test}$, even when the number of samples in $\mathcal{S}_\mathrm{test}$ is small, and, in the case of classification, contains classes that have not been observed in the training set. It is reasonable to assume that for any support/query pair the probability distributions from which the samples have been drawn are the same for the support and the query set. For classification problems, the same holds for the classes that should coincide for the support and query set of one $(\mathcal Q, \mathcal S)$-pair, but not necessarily across all pairs. Since training is performed twice, in the following, we refer to the first stage of training, i.e. on $\mathcal{X}_\mathrm{train}$, as \emph{training}, and the second stage, i.e. on $(\mathcal Q_\mathrm{test}, \mathcal S_\mathrm{test})$ as \emph{adaptation}.

Typically, the models are parameterized by a \emph{task-general} parameter vector $\vtheta$ and a parameter vector $\vvartheta_i$ that is specific to one particular support/query set pair $(\mathcal{Q}_i,\mathcal{S}_i)$. The aim of meta-learning is to use $\mathcal X_\mathrm{train}$ to learn a $\vtheta$ that is as general as possible, such that inferring $\vvartheta_\mathrm{test}$ from a new, unseen support set $\mathcal{S}_\mathrm{test}$ requires as little effort and data as possible (\emph{fast adaptation}).

In \cite{yao2020}, three types of meta-learning approaches have been identified.
\emph{Metric-based} methods learn an embedding space parameterized by $\vtheta$ in which the classes are well separable across all of $X_\mathrm{train}$  w.r.t. some distance measure. An additional, simple proximity-based classifier parameterized by $\vvartheta_i$ is learned jointly for each $i\in\{1,\dots,N_\mathrm{train}\}$. Recent examples of this type of meta-learning models are \cite{yoon2019,oreshkin2018} and \cite{sung2018}.

\emph{Gradient-based} methods minimize a measure of expected non-optimality, such that adaptation requires only few small gradient steps. Prominent examples include \cite{finn2018,yao2020} as well as \cite{grant2018} and \cite{andrychowicz2016}.

The more recent class of \emph{Amortization} methods relies on inference networks that predict the task-specific parameters $\vvartheta_i$ \cite{gordon2018}.

Additionally, to these three classes, remarkably many meta-learning mechanisms rely on \emph{recurrent} models, since meta-learning can be phrased as a sequence-to-sequence problem \cite{ravi2016,mishra2017}

\subsubsection{Self-supervised Learning}
Supervised learning gives us the means to solve tasks for which labels are available in sufficient quantities and variations.
However, the acquisition of the required annotations is usually associated with great effort and high costs.
Meanwhile, a lot of information in the data remains unexploited, labels that are basically free.
In contrast, self-supervised methods try to exploit this untapped potential. Self-supervised learning is an important tool in training skill-invariant models. Since data is not assumed to be consistently labeled, the model is not trained with a specific task in mind.

The general goal is to learn how to encode objects, such as words, images, audio snippets, graphs, etc., into representations that contain the essential information in a condensed form and, thus, can be used to efficiently solve multiple downstream tasks.
To achieve this goal, self-supervised methods formulate tasks for which the labels are automatically provided instead of relying on human-annotated labels.
Typically, the performance on this self-supervised task, often called pretext task, is not important.
The actual goal is that the intermediate representations of the trained model encode high-level semantic information.
The challenge is to design this pretext task in such a way that high-level understanding is necessary to solve it.

One class of self-supervised methods formulates the objective as a prediction task, where a hidden part of the input must be derived from other parts.
This objective comes in many flavors, such as predicting a word in a sentence from context \cite{mikolov2013}, \cite{devlin2018}, inpainting \cite{pathak2016}, colorization \cite{zhang2016} or predicting future frames in a video which will be accessible in subsequent time steps \cite{srivastava2015}.

Another class of methods solves prediction tasks in learned representation spaces;
for example, the relative localization of patches \cite{doersch2015}, \cite{noroozi2016}, the natural orientation of images \cite{gidaris2018}
or the geometric transformation between images \cite{agrawal2015}, \cite{zamir2016}, \cite{zhang2019}.
The potential advantages of the latter techniques are that they have access to the entire input and do not have to learn details at the image level that are irrelevant for understanding image semantics.

In a broader sense, generative models like autoencoders and generative adversarial networks \cite{goodfellow2014} can be considered self-supervised.
However, while the focus of generative models is typically to create realistic and diverse samples, the goal of self-supervised learning is to extract meaningful information from data.
For a broader overview of self-supervised learning we refer the reader to a recent study \cite{jing2019}.

\subsubsection{Metric Learning}
Since systems that are invariant to the data distribution can not rely on solely inferring the training data statistics, distance and metric learning \cite{Weinberger2009}, has gained considerable importance in the last years.

\emph{Deep metric learning} employ deep neural nets to construct in embedding of the data in which the Euclidean distance reflects actual semantic (dis-)similarity between data points. Learning a distance can substitute learning a function $f_\vtheta$ as described in \Secref{sub:skill} in order to become more skill- invariant and distribution-invariant. For instance, instead of learning a function that classifies samples, we can learn a metric, and use a simple distance-based classifier, e.g. $k$ Nearest Neighbors on top, which permits us to neglect the joint probability between the data samples and the labels and rather focus on whether we can capture any semantically meaningful notion of similarity. Additionally, we can employ it for a larger variety of task than mere classification, e.g. clustering or content-based retrieval.

An important approach to metric learning is by \emph{contrastive} embedding. This strategy aims at penalizing pairs of data samples from the same class that are too far apart, as well as pairs of samples from different classes that are too close together. In \cite{Hadsell2006}, this aim is formalized as follows. Let $\vx_i$ and $\vx_j$ be two samples from the training data set $\mathcal{X}_\mathrm{train}$ and $y_{i,j}$ a label that is $0$ if the pair is deemed similar and $1$ otherwise. This yields the loss function
\begin{equation}
L_{\mathrm c}(\vtheta)=\frac{1}{2}\sum_{i,j,i\neq j}(1-y_{i,j})\|f_\vtheta(\vx_i)-f_\vtheta(\vx_j)\|^2+y_{i,j}\max(0, m-\|f_\vtheta(\vx_i)-f_\vtheta(\vx_j)\|)^2,
\end{equation}
where $m>0$ is a threshold value.
Contrastive embedding has been successfully applied to learning similarity of interior design images \cite{Bell2015}.

Alternatively \emph{triplet} loss chooses three samples $\vx^a, \vx^p, \vx^n$ where $\vx^p$ (positive) is assumed to be similar to $\vx_a$ and (anchor) and $\vx^n$ (negative is assumed to be dissimilar from it). Based on these assumptions, the loss function
\begin{equation}
L_{\mathrm{t}}(\vtheta)=\sum_{i}\max(0,\|f_\vtheta(\vx^a_i)-f_\vtheta(\vx^p_i)\|^2-\|f_\vtheta(\vx^a_i)-f_\vtheta(\vx^n_i)\|^2-m),
\end{equation}
where $m$ is again a threshold, is constructed, based on a sufficient number of triplets $\vx^a_i, \vx^p_i, \vx^n_i$. Triplet loss has been successfully employed to face recognition tasks \cite{Schroff2015}, among others.

More recent works propose more sophisticated loss functions, e.g. \emph{Lifted Structured Feature Embedding} \cite{OhSong2016}, \emph{Multi-class $n$-pair loss} \cite{Sohn2016} or angular loss \cite{Wang2017}.

\section{Conclusion}
Knowledge can be expected to play a key role in deep learning and AI developments of the years to come. Many works have investigated the concept of knowledge by emphasizing its interpretation as domain or expert knowledge and developing methods that infuse complementary, problem-specific insights into general-purpose machine learning algorithms. The research questions this type of works tries to answer usually relate to adapting a given model to a specific problem or situation. 

By contrast, many recent trends in machine learning research put the machine learning models themselves at the center of interest, rather than the diverse application scenarios they can be applied to. This shifts the focus from \emph{adaptation} to \emph{adaptability}, and to the challenge of \emph{designing} the models in a way such that the effort involved in adapting them can be minimized.

Motivated by these developments, we conclude that the decisive facet of knowledge in advancing the field is that of \emph{invariance}. Not incidentally, it coincides with definitions from knowledge management. Invariance can refer to different aspects of a machine learning model and, on a low-level, is already a design principle of well-established neural architectures. However, in order to interpret, process, represent or generate knowledge with machine learning, we need to achieve invariance in a broader and more abstract sense. This is a gradual process as there is no clear boundary at which invariance of skill, distribution or syntax is achieved. 

As machine learning models become increasingly invariant, one expects to achieve and enhance the following properties of future industrial and societal developments.
\begin{itemize}
	\item Small data size: One fundamental challenge in real-world application is that
	the size of data available for training an appropriate machine learning model is
	often too small. This is an obstacle researchers and practitioners face all too often, in particular when they need to apply their model to real-world problems where gathering and annotating data is costly and publicly available datasets do not exist.
	By leveraging the advantage of capturing or representing
	intrinsic invariance in data, we expect that models can be learned on small, inconsistent or insufficiently labeled datesets. That way, the bottleneck of industrial applications can be resolved.
	\item Human-like intelligent system: Invariance is crucial in human-centric engineering. The ways humans interact with machines is unique to every user. Systems that interact with humans in a natural and intuitive way thus require the capability to adapt to a large variety of individual traits, such as pronunciation, physical features or design preferences. 
	By becoming increasingly invariant, human-centric systems could reduce their sensitivity to such peculiarities, for instance by permitting models that are trained on a limited set of users to adapt to new human subjects with their own unique habits and preferences.
	\item Multi-purpose intelligent systems: Autonomous systems can be expected to become more versatile and universally applicable in the future, a trend that can already observed today.
	To this end they need the capability to perform different tasks and adapt to diverse situations which can be only achieved with a certain degree of invariance.
\end{itemize}

\bibliographystyle{plain}
\bibliography{nsi}

\begin{thebibliography}{10}

\bibitem{agrawal2015}
Pulkit Agrawal, Joao Carreira, and Jitendra Malik.
\newblock Learning to see by moving.
\newblock In {\em Proceedings of the IEEE international conference on computer
  vision}, pages 37--45, 2015.

\bibitem{andrychowicz2016}
Marcin Andrychowicz, Misha Denil, Sergio Gomez, Matthew~W Hoffman, David Pfau,
  Tom Schaul, Brendan Shillingford, and Nando De~Freitas.
\newblock Learning to learn by gradient descent by gradient descent.
\newblock In {\em Advances in neural information processing systems}, pages
  3981--3989, 2016.

\bibitem{arjovsky2019}
Martin Arjovsky, L{\'e}on Bottou, Ishaan Gulrajani, and David Lopez-Paz.
\newblock Invariant risk minimization.
\newblock {\em arXiv preprint arXiv:1907.02893}, 2019.

\bibitem{bader2005}
Sebastian Bader and Pascal Hitzler.
\newblock Dimensions of neural-symbolic integration - {A} structured survey.
\newblock {\em CoRR}, abs/cs/0511042, 2005.

\bibitem{bahdanau2014}
Dzmitry Bahdanau, Kyunghyun Cho, and Yoshua Bengio.
\newblock Neural machine translation by jointly learning to align and
  translate.
\newblock {\em arXiv preprint arXiv:1409.0473}, 2014.

\bibitem{Bell2015}
Sean Bell and Kavita Bala.
\newblock Learning visual similarity for product design with convolutional
  neural networks.
\newblock {\em ACM transactions on graphics (TOG)}, 34(4):1--10, 2015.

\bibitem{bengio2017}
Yoshua Bengio.
\newblock The consciousness prior.
\newblock {\em arXiv preprint arXiv:1709.08568}, 2017.

\bibitem{besold2017}
Tarek~R. Besold, Artur~S. d'Avila Garcez, Sebastian Bader, Howard Bowman,
  Pedro~M. Domingos, Pascal Hitzler, Kai{-}Uwe K{\"u}hnberger, Lu\'{\i}s~C.
  Lamb, Daniel Lowd, Priscila Machado~Vieira Lima, Leo de~Penning, Gadi Pinkas,
  Hoifung Poon, and Gerson Zaverucha.
\newblock Neural-symbolic learning and reasoning: {A} survey and
  interpretation.
\newblock {\em CoRR}, abs/1711.03902, 2017.

\bibitem{blockeel2011}
H.~Blockeel, K.M. Borgwardt, L.~{De Raedt}, P.~Domingos, K.~Kersting, and
  X.~Yan.
\newblock Guest editorial to the special issue on inductive logic programming,
  mining and learning in graphs and statistical relational learning.
\newblock {\em Machine Learning}, 83:133--135, 2011.

\bibitem{bojchevski2018}
Aleksandar Bojchevski, Oleksandr Shchur, Daniel Z{\"u}gner, and Stephan
  G{\"u}nnemann.
\newblock Netgan: Generating graphs via random walks.
\newblock {\em arXiv preprint arXiv:1803.00816}, 2018.

\bibitem{bruna2013}
Joan Bruna, Wojciech Zaremba, Arthur Szlam, and Yann LeCun.
\newblock Spectral networks and locally connected networks on graphs.
\newblock {\em arXiv preprint arXiv:1312.6203}, 2013.

\bibitem{chollet2019}
Fran{\c c}ois Chollet.
\newblock The measure of intelligence.
\newblock {\em arXiv preprint arXiv:1911.01547}, 2019.

\bibitem{dai2018}
Hanjun Dai, Zornitsa Kozareva, Bo~Dai, Alex Smola, and Le~Song.
\newblock Learning steady-states of iterative algorithms over graphs.
\newblock In {\em International conference on machine learning}, pages
  1106--1114, 2018.

\bibitem{davenport2000}
T.H. Davenport and L.~Prusak.
\newblock {\em Working Knowledge: How Organizations Manage what They Know}.
\newblock EBSCO eBook Collection. Harvard Business School Press, 2000.

\bibitem{garcez2007}
A.~d'Avila Garcez, L.~C. Lamb, and D.~M. Gabbay.
\newblock Connec- tionist modal logic: Representing modalities in neural
  networks.
\newblock {\em Theoretical Computer Science}, 371:34--53, 2007.

\bibitem{penning2010}
L~de~Penning, A~d’Avila Garcez, L.~C. Lamb, and J.~J. Meyer.
\newblock An integrated neural-symbolic cognitive agent architecture for
  training and assessment in simulators.
\newblock {\em Proceedings of the Twenty-Second International Joint Conference
  on Artificial Intelligence, held at AAAI-2010}, pages 1653--1658, 2010.

\bibitem{penning2011}
L~de~Penning, A~d’Avila Garcez, L.~C. Lamb, and J.~J. Meyer.
\newblock A neural-symbolic cognitive agent for online learning and reasoning.
\newblock In {\em 22nd International Joint Conference on Artificial
  Intelligence}, 2011.

\bibitem{devlin2018}
Jacob Devlin, Ming-Wei Chang, Kenton Lee, and Kristina Toutanova.
\newblock Bert: Pre-training of deep bidirectional transformers for language
  understanding.
\newblock {\em arXiv preprint arXiv:1810.04805}, 2018.

\bibitem{diehl2019}
Frederik Diehl, Thomas Brunner, Michael~Truong Le, and Alois Knoll.
\newblock Towards graph pooling by edge contraction.
\newblock In {\em ICML 2019 Workshop on Learning and Reasoning with
  Graph-Structured Data}, 2019.

\bibitem{doersch2015}
Carl Doersch, Abhinav Gupta, and Alexei~A Efros.
\newblock Unsupervised visual representation learning by context prediction.
\newblock In {\em Proceedings of the IEEE International Conference on Computer
  Vision}, pages 1422--1430, 2015.

\bibitem{finn2017}
Chelsea Finn, Pieter Abbeel, and Sergey Levine.
\newblock Model-agnostic meta-learning for fast adaptation of deep networks.
\newblock In {\em Proceedings of the 34th International Conference on Machine
  Learning-Volume 70}, pages 1126--1135. JMLR. org, 2017.

\bibitem{finn2018}
Chelsea Finn, Kelvin Xu, and Sergey Levine.
\newblock Probabilistic model-agnostic meta-learning.
\newblock In {\em Advances in Neural Information Processing Systems}, pages
  9516--9527, 2018.

\bibitem{flach2012}
Peter Flach.
\newblock {\em Machine learning: the art and science of algorithms that make
  sense of data}.
\newblock Cambridge University Press, 2012.

\bibitem{garcez2009}
A.~Garcez, L.~C. Lamb, and D.M. Gabbay.
\newblock {\em Neural-Symbolic Cognitive Reasoning}.
\newblock Cognitive Technologies. Springer, 2009.

\bibitem{garcez1999}
A~Garcez and G~Zaverucha.
\newblock The connectionist inductive learning and logic programming system.
\newblock {\em Applied Intelligence}, 11:59--77, 1999.

\bibitem{garcez2015}
Artur~d'Avila Garcez, Tarek~R. Besold, Luc~de Raedt, Peter F{\"o}ldiak, Pascal
  Hitzler, Thomas Icard, Kai-Uwe K{\"u}hnberger, Luis~C. Lamb, Risto
  Miikkulainen, and Daniel~L. Silver.
\newblock Neural-symbolic learning and reasoning: Contributions and challenges.
\newblock {\em Knowledge Representation and Reasoning}, pages 18--21, 2015.

\bibitem{gidaris2018}
Spyros Gidaris, Praveer Singh, and Nikos Komodakis.
\newblock Unsupervised representation learning by predicting image rotations.
\newblock {\em arXiv preprint arXiv:1803.07728}, 2018.

\bibitem{goodfellow2014}
Ian Goodfellow, Jean Pouget-Abadie, Mehdi Mirza, Bing Xu, David Warde-Farley,
  Sherjil Ozair, Aaron Courville, and Yoshua Bengio.
\newblock Generative adversarial nets.
\newblock In {\em Advances in neural information processing systems}, pages
  2672--2680, 2014.

\bibitem{gordon2014}
Andrew~D Gordon, Thomas~A Henzinger, Aditya~V Nori, and Sriram~K Rajamani.
\newblock Probabilistic programming.
\newblock In {\em Proceedings of the on Future of Software Engineering}, pages
  167--181. ACM, 2014.

\bibitem{gordon2018}
Jonathan Gordon, John Bronskill, Matthias Bauer, Sebastian Nowozin, and
  Richard~E Turner.
\newblock Meta-learning probabilistic inference for prediction.
\newblock {\em arXiv preprint arXiv:1805.09921}, 2018.

\bibitem{grant2018}
Erin Grant, Chelsea Finn, Sergey Levine, Trevor Darrell, and Thomas Griffiths.
\newblock Recasting gradient-based meta-learning as hierarchical bayes.
\newblock {\em arXiv preprint arXiv:1801.08930}, 2018.

\bibitem{Greenfeld2019}
Daniel Greenfeld and Uri Shalit.
\newblock Robust learning with the hilbert-schmidt independence criterion.
\newblock {\em arXiv preprint arXiv:1910.00270}, 2019.

\bibitem{Hadsell2006}
Raia Hadsell, Sumit Chopra, and Yann LeCun.
\newblock Dimensionality reduction by learning an invariant mapping.
\newblock In {\em 2006 IEEE Computer Society Conference on Computer Vision and
  Pattern Recognition (CVPR'06)}, volume~2, pages 1735--1742. IEEE, 2006.

\bibitem{harel2001}
D~Harel, D~Kozen, and J~Tiuryn.
\newblock Dynamic logic.
\newblock {\em SIGACT News}, 32:66--69, 2001.

\bibitem{hinton2011}
Geoffrey~E Hinton, Alex Krizhevsky, and Sida~D Wang.
\newblock Transforming auto-encoders.
\newblock In {\em International conference on artificial neural networks},
  pages 44--51. Springer, 2011.

\bibitem{dkiw2015}
WikiCommons Image: Longlivetheux / CC BY-SA
  (https://creativecommons.org/licenses/by sa/4.0).

\bibitem{hu2018squeeze}
Jie Hu, Li~Shen, and Gang Sun.
\newblock {Squeeze-and-excitation networks}.
\newblock In {\em Proceedings of the IEEE conference on computer vision and
  pattern recognition}, pages 7132--7141, 2018.

\bibitem{jing2019}
Longlong Jing and Yingli Tian.
\newblock Self-supervised visual feature learning with deep neural networks: A
  survey.
\newblock {\em arXiv preprint arXiv:1902.06162}, 2019.

\bibitem{kahneman2011}
Daniel Kahneman.
\newblock {\em Thinking, fast and slow}.
\newblock Macmillan, 2011.

\bibitem{kim2017interpretable}
Jinkyu Kim and John Canny.
\newblock {Interpretable learning for self-driving cars by visualizing causal
  attention}.
\newblock In {\em Proceedings of the IEEE international conference on computer
  vision}, pages 2942--2950, 2017.

\bibitem{kipf2016}
Thomas~N Kipf and Max Welling.
\newblock Semi-supervised classification with graph convolutional networks.
\newblock {\em arXiv preprint arXiv:1609.02907}, 2016.

\bibitem{koller2009}
D~Koller and N~Friedman.
\newblock {\em Probabilistic Graphical Models: Principles and Techniques}.
\newblock MIT Press, 2009.

\bibitem{kondor2018}
Risi Kondor and Shubhendu Trivedi.
\newblock On the generalization of equivariance and convolution in neural
  networks to the action of compact groups.
\newblock {\em arXiv preprint arXiv:1802.03690}, 2018.

\bibitem{kosiorek2019}
Adam Kosiorek, Sara Sabour, Yee~Whye Teh, and Geoffrey~E Hinton.
\newblock Stacked capsule autoencoders.
\newblock In {\em Advances in Neural Information Processing Systems}, pages
  15486--15496, 2019.

\bibitem{krotschz2013}
M.~Kr{\"o}tzsch, S.~Rudolph, and P.~Hitzler.
\newblock Complexity of horn description logics.
\newblock {\em ACM Trans. Comput. Logic}, 44, 2013.

\bibitem{LeCun2020}
Yann LeCun, Yoshua Bengio, and Geoffrey Hinton.
\newblock Keynote Talk at AAAI, 2020.

\bibitem{lee2009}
Honglak Lee, Roger Grosse, Rajesh Ranganath, and Andrew~Y Ng.
\newblock Convolutional deep belief networks for scalable unsupervised learning
  of hierarchical representations.
\newblock In {\em Proceedings of the 26th annual international conference on
  machine learning}, pages 609--616, 2009.

\bibitem{lee2018}
Juho Lee, Yoonho Lee, Jungtaek Kim, Adam~R Kosiorek, Seungjin Choi, and
  Yee~Whye Teh.
\newblock Set transformer: A framework for attention-based
  permutation-invariant neural networks.
\newblock {\em arXiv preprint arXiv:1810.00825}, 2018.

\bibitem{li2018}
Ruoyu Li, Sheng Wang, Feiyun Zhu, and Junzhou Huang.
\newblock Adaptive graph convolutional neural networks.
\newblock In {\em Thirty-second AAAI conference on artificial intelligence},
  2018.

\bibitem{li2018a}
Yangyan Li, Rui Bu, Mingchao Sun, Wei Wu, Xinhan Di, and Baoquan Chen.
\newblock Pointcnn: Convolution on x-transformed points.
\newblock In {\em Advances in neural information processing systems}, pages
  820--830, 2018.

\bibitem{li2015}
Yujia Li, Daniel Tarlow, Marc Brockschmidt, and Richard Zemel.
\newblock Gated graph sequence neural networks.
\newblock {\em arXiv preprint arXiv:1511.05493}, 2015.

\bibitem{lifschitz2002}
V.~Lifschitz.
\newblock Answer set programming and plan generation.
\newblock {\em Artificial Intelligence}, 138:39--54, 2002.

\bibitem{Mallat2016}
St{\'e}phane Mallat.
\newblock Understanding deep convolutional networks.
\newblock {\em Philosophical Transactions of the Royal Society A: Mathematical,
  Physical and Engineering Sciences}, 374(2065):20150203, 2016.

\bibitem{marcus2020}
Gary Marcus.
\newblock The next decade in ai: four steps towards robust artificial
  intelligence.
\newblock {\em arXiv preprint arXiv:2002.06177}, 2020.

\bibitem{mikolov2013}
Tomas Mikolov, Kai Chen, Greg Corrado, and Jeffrey Dean.
\newblock Efficient estimation of word representations in vector space.
\newblock {\em arXiv preprint arXiv:1301.3781}, 2013.

\bibitem{mishra2017}
Nikhil Mishra, Mostafa Rohaninejad, Xi~Chen, and Pieter Abbeel.
\newblock A simple neural attentive meta-learner.
\newblock {\em arXiv preprint arXiv:1707.03141}, 2017.

\bibitem{monti2017}
Federico Monti, Davide Boscaini, Jonathan Masci, Emanuele Rodola, Jan Svoboda,
  and Michael~M Bronstein.
\newblock Geometric deep learning on graphs and manifolds using mixture model
  cnns.
\newblock In {\em Proceedings of the IEEE Conference on Computer Vision and
  Pattern Recognition}, pages 5115--5124, 2017.

\bibitem{noroozi2016}
Mehdi Noroozi and Paolo Favaro.
\newblock Unsupervised learning of visual representations by solving jigsaw
  puzzles.
\newblock In {\em European Conference on Computer Vision}, pages 69--84.
  Springer, 2016.

\bibitem{OhSong2016}
Hyun Oh~Song, Yu~Xiang, Stefanie Jegelka, and Silvio Savarese.
\newblock Deep metric learning via lifted structured feature embedding.
\newblock In {\em Proceedings of the IEEE conference on computer vision and
  pattern recognition}, pages 4004--4012, 2016.

\bibitem{oreshkin2018}
Boris Oreshkin, Pau~Rodr{\'\i}guez L{\'o}pez, and Alexandre Lacoste.
\newblock Tadam: Task dependent adaptive metric for improved few-shot learning.
\newblock In {\em Advances in Neural Information Processing Systems}, pages
  721--731, 2018.

\bibitem{pang2019mask}
Yanwei Pang, Jin Xie, Muhammad~Haris Khan, Rao~Muhammad Anwer, Fahad~Shahbaz
  Khan, and Ling Shao.
\newblock Mask-guided attention network for occluded pedestrian detection.
\newblock In {\em Proceedings of the IEEE International Conference on Computer
  Vision}, pages 4967--4975, 2019.

\bibitem{pathak2016}
Deepak Pathak, Philipp Krahenbuhl, Jeff Donahue, Trevor Darrell, and Alexei~A
  Efros.
\newblock Context encoders: Feature learning by inpainting.
\newblock In {\em Proceedings of the IEEE conference on computer vision and
  pattern recognition}, pages 2536--2544, 2016.

\bibitem{perozzi2014}
Bryan Perozzi, Rami Al-Rfou, and Steven Skiena.
\newblock Deepwalk: Online learning of social representations.
\newblock In {\em Proceedings of the 20th ACM SIGKDD international conference
  on Knowledge discovery and data mining}, pages 701--710, 2014.

\bibitem{pnueli1977}
A~Pnueli.
\newblock The temporal logic of programs.
\newblock {\em 18th Annual Symposium on Foundations of Computer Science}, pages
  46--57, 1977.

\bibitem{ramachandran2019stand}
Prajit Ramachandran, Niki Parmar, Ashish Vaswani, Irwan Bello, Anselm Levskaya,
  and Jonathon Shlens.
\newblock Stand-alone self-attention in vision models.
\newblock {\em arXiv preprint arXiv:1906.05909}, 2019.

\bibitem{ravanbakhsh2017}
Siamak Ravanbakhsh, Jeff Schneider, and Barnabas Poczos.
\newblock Equivariance through parameter-sharing.
\newblock In {\em Proceedings of the 34th International Conference on Machine
  Learning-Volume 70}, pages 2892--2901. JMLR. org, 2017.

\bibitem{ravi2016}
Sachin Ravi and Hugo Larochelle.
\newblock Optimization as a model for few-shot learning.
\newblock 2016.

\bibitem{rowley2008}
J.E. Rowley and R.J. Hartley.
\newblock {\em Organizing Knowledge: An Introduction to Managing Access to
  Information}.
\newblock Ashgate, 2008.

\bibitem{rowley2007}
Jennifer Rowley.
\newblock The wisdom hierarchy: representations of the dikw hierarchy.
\newblock {\em Journal of Information Science}, 33(2):163--180, 2007.

\bibitem{Schroff2015}
Florian Schroff, Dmitry Kalenichenko, and James Philbin.
\newblock Facenet: A unified embedding for face recognition and clustering.
\newblock In {\em Proceedings of the IEEE conference on computer vision and
  pattern recognition}, pages 815--823, 2015.

\bibitem{Sohn2016}
Kihyuk Sohn.
\newblock Improved deep metric learning with multi-class n-pair loss objective.
\newblock In {\em Advances in neural information processing systems}, pages
  1857--1865, 2016.

\bibitem{srivastava2015}
Nitish Srivastava, Elman Mansimov, and Ruslan Salakhudinov.
\newblock Unsupervised learning of video representations using lstms.
\newblock In {\em International conference on machine learning}, pages
  843--852, 2015.

\bibitem{sung2018}
Flood Sung, Yongxin Yang, Li~Zhang, Tao Xiang, Philip~HS Torr, and Timothy~M
  Hospedales.
\newblock Learning to compare: Relation network for few-shot learning.
\newblock In {\em Proceedings of the IEEE Conference on Computer Vision and
  Pattern Recognition}, pages 1199--1208, 2018.

\bibitem{ulyanov2018}
Dmitry Ulyanov, Andrea Vedaldi, and Victor Lempitsky.
\newblock Deep image prior.
\newblock In {\em Proceedings of the IEEE Conference on Computer Vision and
  Pattern Recognition}, pages 9446--9454, 2018.

\bibitem{dalen2002}
D~Van~Dalen.
\newblock Intuitionistic logic.
\newblock In {\em In Handbook of philosophical logic}, pages 1--114. Springer,
  2002.

\bibitem{vaswani2017}
Ashish Vaswani, Noam Shazeer, Niki Parmar, Jakob Uszkoreit, Llion Jones,
  Aidan~N Gomez, {\L}ukasz Kaiser, and Illia Polosukhin.
\newblock Attention is all you need.
\newblock In {\em Advances in neural information processing systems}, pages
  5998--6008, 2017.

\bibitem{velivckovic2017}
Petar Veli{\v{c}}kovi{\'c}, Guillem Cucurull, Arantxa Casanova, Adriana Romero,
  Pietro Lio, and Yoshua Bengio.
\newblock Graph attention networks.
\newblock {\em arXiv preprint arXiv:1710.10903}, 2017.

\bibitem{velivckovic2018}
Petar Veli{\v{c}}kovi{\'c}, William Fedus, William~L Hamilton, Pietro Li{\`o},
  Yoshua Bengio, and R~Devon Hjelm.
\newblock Deep graph infomax.
\newblock {\em arXiv preprint arXiv:1809.10341}, 2018.

\bibitem{vinyals2015}
Oriol Vinyals, Samy Bengio, and Manjunath Kudlur.
\newblock Order matters: Sequence to sequence for sets.
\newblock {\em arXiv preprint arXiv:1511.06391}, 2015.

\bibitem{vinyals2016}
Oriol Vinyals, Charles Blundell, Timothy Lillicrap, Daan Wierstra, et~al.
\newblock Matching networks for one shot learning.
\newblock In {\em Advances in neural information processing systems}, pages
  3630--3638, 2016.

\bibitem{VonRueden2019}
Laura Von~Rueden, Sebastian Mayer, Jochen Garcke, Christian Bauckhage, and
  Jannis Schuecker.
\newblock Informed machine learning--towards a taxonomy of explicit integration
  of knowledge into machine learning.
\newblock {\em Learning}, 18:19--20, 2019.

\bibitem{Wang2017}
Jian Wang, Feng Zhou, Shilei Wen, Xiao Liu, and Yuanqing Lin.
\newblock Deep metric learning with angular loss.
\newblock In {\em Proceedings of the IEEE International Conference on Computer
  Vision}, pages 2593--2601, 2017.

\bibitem{Wang2018}
Xiaolong Wang, Ross Girshick, Abhinav Gupta, and Kaiming He.
\newblock {Non-local Neural Networks}.
\newblock {\em Proceedings of the IEEE Computer Society Conference on Computer
  Vision and Pattern Recognition}, pages 7794--7803, 2018.

\bibitem{Weinberger2009}
Kilian~Q Weinberger and Lawrence~K Saul.
\newblock Distance metric learning for large margin nearest neighbor
  classification.
\newblock {\em Journal of Machine Learning Research}, 10(2), 2009.

\bibitem{wu2020}
Zonghan Wu, Shirui Pan, Fengwen Chen, Guodong Long, Chengqi Zhang, and S~Yu
  Philip.
\newblock A comprehensive survey on graph neural networks.
\newblock {\em IEEE Transactions on Neural Networks and Learning Systems},
  2020.

\bibitem{Xu:2015:SAT:3045118.3045336}
Kelvin Xu, Jimmy~Lei Ba, Ryan Kiros, Kyunghyun Cho, Aaron Courville, Ruslan
  Salakhutdinov, Richard~S Zemel, and Yoshua Bengio.
\newblock {Show, Attend and Tell: Neural Image Caption Generation with Visual
  Attention}.
\newblock In {\em Proceedings of the 32Nd International Conference on
  International Conference on Machine Learning - Volume 37}, ICML'15, pages
  2048--2057. JMLR.org, 2015.

\bibitem{xu2018}
Keyulu Xu, Weihua Hu, Jure Leskovec, and Stefanie Jegelka.
\newblock How powerful are graph neural networks?
\newblock {\em arXiv preprint arXiv:1810.00826}, 2018.

\bibitem{yao2020}
Huaxiu Yao, Xian Wu, Zhiqiang Tao, Yaliang Li, Bolin Ding, Ruirui Li, and
  Zhenhui Li.
\newblock Automated relational meta-learning.
\newblock {\em arXiv preprint arXiv:2001.00745}, 2020.

\bibitem{yoon2019}
Sung~Whan Yoon, Jun Seo, and Jaekyun Moon.
\newblock Tapnet: Neural network augmented with task-adaptive projection for
  few-shot learning.
\newblock {\em arXiv preprint arXiv:1905.06549}, 2019.

\bibitem{zaheer2017}
Manzil Zaheer, Satwik Kottur, Siamak Ravanbakhsh, Barnabas Poczos, Russ~R
  Salakhutdinov, and Alexander~J Smola.
\newblock Deep sets.
\newblock In {\em Advances in neural information processing systems}, pages
  3391--3401, 2017.

\bibitem{zamir2016}
Amir~R Zamir, Tilman Wekel, Pulkit Agrawal, Colin Wei, Jitendra Malik, and
  Silvio Savarese.
\newblock Generic 3d representation via pose estimation and matching.
\newblock In {\em European Conference on Computer Vision}, pages 535--553.
  Springer, 2016.

\bibitem{zhang2019}
Liheng Zhang, Guo-Jun Qi, Liqiang Wang, and Jiebo Luo.
\newblock Aet vs. aed: Unsupervised representation learning by auto-encoding
  transformations rather than data.
\newblock In {\em Proceedings of the IEEE Conference on Computer Vision and
  Pattern Recognition}, pages 2547--2555, 2019.

\bibitem{zhang2016}
Richard Zhang, Phillip Isola, and Alexei~A Efros.
\newblock Colorful image colorization.
\newblock In {\em European conference on computer vision}, pages 649--666.
  Springer, 2016.

\bibitem{zhang2018}
Shanshan Zhang, Jian Yang, and Bernt Schiele.
\newblock {Occluded pedestrian detection through guided attention in CNNs}.
\newblock In {\em Proceedings of the IEEE Conference on Computer Vision and
  Pattern Recognition}, pages 6995--7003, 2018.

\end{thebibliography}
\end{document}